\documentclass[letterpaper, 10 pt, journal, twoside]{ieeetran}

\IEEEoverridecommandlockouts                              

\usepackage{amssymb}
\usepackage{amsmath,amsfonts}
\usepackage{algorithmic}
\usepackage{algorithm}
\usepackage{array}
\usepackage[caption=false,font=normalsize,labelfont=sf,textfont=sf]{subfig}
\usepackage{textcomp}
\usepackage{stfloats}
\usepackage{url}
\usepackage{verbatim}
\usepackage{graphicx}
\usepackage{cite}
\usepackage{multirow}
\usepackage{booktabs}
\usepackage{hyperref}
\usepackage{xcolor}
\usepackage{xcolor}
\usepackage{soul}
\newcolumntype{C}[1]{>{\centering\arraybackslash}m{#1}}
\usepackage{makecell}

\title{
Attention-Based Neural-Augmented Kalman Filter for Legged Robot State Estimation
}

\author{Seokju Lee\href{https://orcid.org/0009-0009-7620-2403}{\includegraphics[scale=0.017]{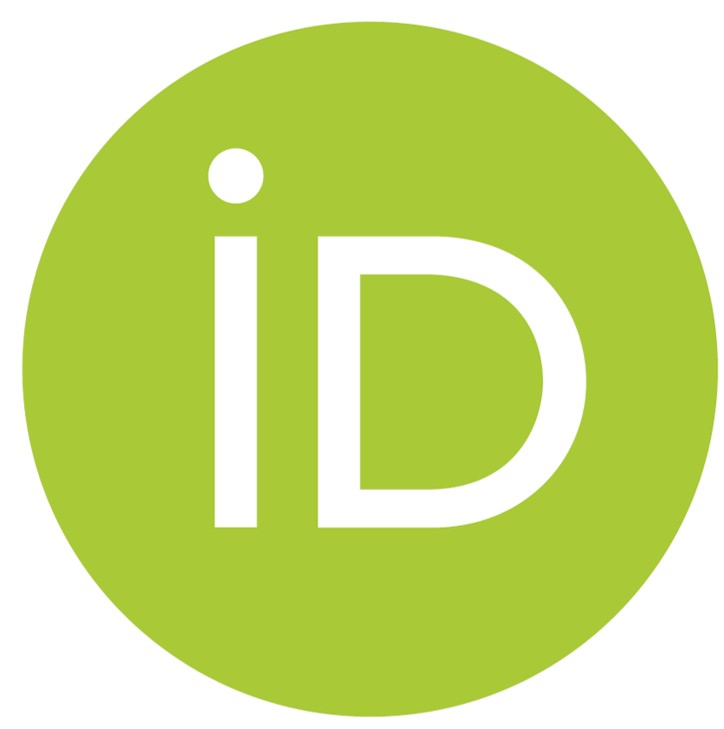}},~\IEEEmembership{Graduate Student Member,~IEEE,} and Kyung-Soo Kim\href{https://orcid.org/0000-0003-4856-1096}{\includegraphics[scale=0.017]{orcid_mark.jpg}},~\IEEEmembership{Member,~IEEE}
\thanks{Manuscript received: October 4, 2025; Revised December 21, 2025; Accepted January 24, 2026.}
\thanks{This paper was recommended for publication by Editor Abderrahmane Kheddar upon evaluation of the Associate Editor and Reviewers’ comments, and is based on the author’s Master’s thesis submitted to the Korea Advanced Institute of Science and Technology (KAIST) in February 2025.} 
\thanks{This work was supported by the BK21 FOUR Program of the National Research Foundation Korea (NRF) grant funded by the Ministry of Education (MOE). \textit{(Corresponding author: Kyung-Soo Kim)}}
\thanks{The authors are with the Mechatronics, Systems and Control Lab (MSC Lab), Department of Mechanical Engineering, KAIST, Yuseong-gu, Daejeon 34141, Republic of Korea (e-mail: \{\texttt{dltjrwn0322, kyungsookim}\}@kaist.ac.kr)}
\thanks{Project Page: \url{https://seokju-lee.github.io/attennkf}}
\thanks{Digital Object Identifier (DOI): see top of this page.}
}%

\begin{document}
\markboth{IEEE Robotics and Automation Letters. Preprint Version. Accepted January, 2026}
{Lee \MakeLowercase{\textit{et al.}}: Attention-Based Neural-Augmented Kalman Filter for Legged Robot State Estimation}

\maketitle
\begin{abstract}
In this letter, we propose an Attention-Based Neural-Augmented Kalman Filter (AttenNKF) for state estimation in legged robots. Foot slip is a major source of estimation error: when slip occurs, kinematic measurements violate the no-slip assumption and inject bias during the update step. Our objective is to estimate this slip-induced error and compensate for it. To this end, we augment an Invariant Extended Kalman Filter (InEKF) with a neural compensator that uses an attention mechanism to infer error conditioned on foot-slip severity and then applies this estimate as a post-update compensation to the InEKF state (i.e., after the filter update). The compensator is trained in a latent space, which aims to reduce sensitivity to raw input scales and encourages structured slip-conditioned compensations, while preserving the InEKF recursion. Experiments demonstrate improved performance compared to existing legged-robot state estimators, particularly under slip-prone conditions.
\end{abstract}

\begin{IEEEkeywords}
Legged robots, proprioceptive state estimation, attention mechanisms
\end{IEEEkeywords}

\section{INTRODUCTION}
\IEEEPARstart{R}{ecently}, there has been substantial research on legged robots capable of traversing diverse terrains \cite{lee2020learning, miki2022learning, choi2023learning, grandia2023perceptive}. To precisely control such robots, it is common to design and optimize control algorithms based on the robot’s pose, velocity, and contact state. In particular, agile and robust locomotion—such as dynamic walking on uneven or low-friction terrain—requires that these quantities be estimated accurately in real time. However, the robot’s internal state cannot be measured directly and must instead be inferred from noisy, indirect onboard sensor readings. As a result, estimating the desired robot state from available measurements becomes a core problem that directly limits the achievable control performance.

In robotic state estimation, approaches based on exteroceptive sensors such as LiDAR, cameras, and radar are widely used. However, camera- and LiDAR-based methods exhibit pronounced performance degradation under adverse weather conditions such as rain, snow, and fog \cite{teufel2023enhancing, zhang2023perception, park2025simulation}, while radar remains robust in such conditions but suffers from limited angular resolution and clutter, which restrict its ability to capture fine-grained terrain detail \cite{han20234d, han2024denserradar}. To mitigate these issues, Bijelic \textit{et al}.~\cite{bijelic2020seeing} propose a multimodal perception framework that combines camera, LiDAR, and radar, but this approach is constrained by the difficulty of collecting large-scale training data in extreme environments. For these reasons, state estimators that rely solely on proprioceptive sensors are essential for legged robots that must operate robustly across a wide range of environments, including adverse weather.

A typical legged robot state estimator integrates the linear acceleration, angular velocity from Inertial Measurement Unit (IMU), and joint positions/velocities obtained from joint encoders to predict the nominal state (predict step), and then updates the posterior state (update step) by computing the residual between the measured and predicted quantities using forward kinematics of the foot and applying the Kalman gain. In this process, the foot kinematics constrain the world frame only for feet in contact, and the corresponding contact points are typically modeled as quasi-static fixed points.

Bloesch \textit{et al}.~\cite{Bloesch-RSS-12} treated small foot slip in such leg-kinematics–based Extended Kalman Filter (EKF) as process noise on the foot position and absorbed it via a random walk. Subsequently, Bloesch \textit{et al}.~\cite{bloesch2013state} proposed to treat measurements that indicate large slip as outliers and to discard the leg-kinematic update at those time steps. Hartley \textit{et al}.~\cite{hartley2020contact} introduced the Invariant EKF (InEKF), which defines the state on a Lie group and achieves faster convergence than a quaternion-based EKF (QEKF). Later, Kim \textit{et al}.~\cite{kim2021legged} and Yoon \textit{et al}.~\cite{yoon2023invariant} incorporated foot-velocity–based slip detection and covariance inflation to reduce the measurement confidence during slip, yet all these methods handle slip only indirectly by tuning the measurement covariance or skipping updates when slip is detected. As a result, they cannot explicitly estimate and directly compensate the bias induced by slip. Meanwhile, Camurri \textit{et al.}~\cite{sh-ch12-proprio} also point out that slip and contact detection are critical factors in legged robot odometry, and report that failing to handle these effects appropriately can lead to severe drift.

Model-based filters augmented with neural networks have also been explored. Lin \textit{et al}.~\cite{lin2022legged} learned a convolutional neural network (CNN)–based contact estimator from IMU and joint data and used its output as the contact measurement in an InEKF, achieving improved performance over a Ground Reaction Force (GRF)–based InEKF. Buchanan \textit{et al}.~\cite{buchanan2022learning} proposed EKF-IKD (EKF with inertial, kinematic, and learned displacement), in which a neural network predicts both displacement and covariance that are then integrated into an EKF. Youm \textit{et al}.~\cite{youm2025legged} introduced a Neural Measurement Network (NMN) that outputs contact probability and estimated foot velocity, leading to improved performance across various terrains. Nevertheless, accumulated slip-induced errors, particularly in the $z$-direction position, are still observed. 

In related work, Lee \textit{et al}.~\cite{lee2025legged} proposed the Invariant Neural-Augmented Kalman Filter (InNKF), which introduces a Neural Compensator (NC) that applies an additional compensation to the state estimated by the InEKF. InNKF directly regresses the error using a single Temporal Convolutional Network (TCN) that takes as input a history $\mathcal{H}_t$ of the InEKF posterior, contact information, and sensor measurements. This history-to-error mapping improves estimation performance, but it does not explicitly model slip as a conditioning variable and combines heterogeneous signals with different scales and noise statistics. As a result, the learned compensator behavior can be harder to interpret and tune consistently across different slip regimes. 

This work explicitly addresses these limitations by reformulating the NC as a slip-conditioned function $\bar{\mathbf{e}}_t = f(\bar{\mathbf{x}}_t \,|\, \mathrm{slip}_t)$, treating slip not as a feature but as a context that determines the compensation structure. Since the required compensation depends on the slip regime, a data-dependent compensation pattern is required that can selectively emphasize or suppress elements of the InEKF representation according to slip. We adopt a cross-attention~\cite{vaswani2017attention} module because our goal is to treat slip as a context that modulates how the InEKF latent should be used for compensation, rather than as just another concatenated input feature. The slip latent provides the query, and the InEKF latent provides keys/values, producing a slip-dependent compensation over the history used by the decoder. While other parametrizations of slip-conditioned compensation are conceivable (e.g., concatenation followed by an MLP), our ablation study in Section IV-C supports the benefit of the proposed cross-attention structure under slip. Under higher foot slip levels, the attention weights can shift across the InEKF history, enabling slip-conditioned reweighting of the latent features. When the slip regime changes, the attention distribution changes accordingly, producing slip-dependent compensations compared to slip-agnostic conditioning. The resulting estimator, obtained by augmenting the InEKF with this module, is termed the Attention-based Neural-Augmented Kalman Filter (AttenNKF). Instead of concatenating slip as an input feature, we encode slip separately and condition the compensation using cross-attention over the InEKF history. The main contributions are summarized as follows:
\begin{itemize}
  \item We define a continuous foot slip level using foot velocity and estimated contact state, and analyze the correlation between slip level and estimation error, thereby justifying the use of slip as an explicit context variable.
  \item We propose an NC conditioned on slip and design an Attention-Based Neural-Augmented Kalman Filter (AttenNKF) in which cross-attention between the InEKF and slip latents yields slip-regime–dependent structures for reducing the slip-induced error.
  \item We evaluate AttenNKF on indoor terrains consisting of gravel, a low-friction Teflon sheet, and stairs, as well as on an outdoor trajectory of about 100\,m, and experimentally demonstrate consistent improvements in Relative Error (RE) over Slip Rejection (SR), Learned Contact (LC), and InNKF.
\end{itemize}

The remainder of this paper is organized as follows. Section \ref{section:slip_define} defines the foot slip level, and Section \ref{section:attennkf} provides a detailed description of the training methodology and architecture of AttenNKF. Section \ref{section:experiments} presents the experimental setup and results, while Section \ref{section:conclusion} offers conclusions and future works.
\section{Continuous Foot Slip Level}
\label{section:slip_define}
In this section, we define a continuous foot slip level using the estimated contact state and foot velocity, and then analyze how the resulting slip level relates to state estimation error. Aiming for a general state estimator that does not rely on specific hardware such as force/torque sensors, we adopt the CNN-based contact estimator of Lin \textit{et al}.~\cite{lin2022legged}, which uses only IMU and joint data, and reports about 97\% contact-state classification accuracy across various terrains, to obtain a binary stance/swing contact label for each foot at each time step. For time steps classified as stance, we define the foot slip level by mapping the magnitude of the forward-kinematics-based foot velocity in the world frame through a sigmoid function, yielding a continuous value in $[0,1]$.

\subsection{Foot Slip Level Function}
Based on the estimated contact state obtained from the contact estimator, the indices of the feet in contact with the ground are identified. In the case of static contact, the foot velocity in the world frame, denoted as $\dot{\textbf{d}}^{w}$, should theoretically be close to zero. This foot velocity $\dot{\textbf{d}}^{\text{w}}$ is computed from estimated base state and joint states via forward kinematics, as in \cite{kim2021legged}.

For each time step and each foot $i$, we define a continuous slip-severity indicator by mapping the speed $\|\dot{\textbf{d}}_i^{w}\|$ through a sigmoid function when the foot is classified as in contact. This produces a bounded value in $[0,1]$ that orders the instantaneous degree of slip, rather than estimating slip displacement:
\begin{equation}
lv_i(\dot{\textbf{d}}_i^\text{w}; C_i) = \mathbf{1}_{\{C_i = \text{true}\}} \cdot \frac{1}{1 + \exp\left(-k(\|\dot{\textbf{d}}_i^\text{w}\| - v^{th})\right)}
\label{eq:foot_slip}
\end{equation}
where $\mathbf{1}$ denotes the indicator function, $C_i$ represents the $i$-th foot's contact state, and $k$ and $v^{th}$ are empirically selected parameters denoting the sigmoid steepness and foot-velocity threshold, respectively.

The motivation for using a velocity-based slip signal is twofold: (i) the foot speed during stance directly reflects violations of the (explicit or implicit) zero-velocity contact assumption underlying kinematics-based estimators, and (ii) the signal is available instantaneously and is therefore suitable as a context input for the post-update compensation produced by the neural compensator. Accordingly, we use $lv_i$ only as a severity-ordered context cue, not as a direct proxy for true slip distance. While errors in the contact estimator can introduce noise into the slip level, this uncertainty did not empirically compromise estimation performance in our experiments, since $lv_i$ is not used as a direct measurement or supervision signal of the filter. However, if contact misclassification persists over many consecutive steps, the resulting slip level can become misleading, potentially limiting the compensator's ability to fully correct slip-induced errors.
\subsection{Correlation of Foot Slip and State Estimation Error}
\begin{figure}
    \centering
    \includegraphics[width=0.83\linewidth]{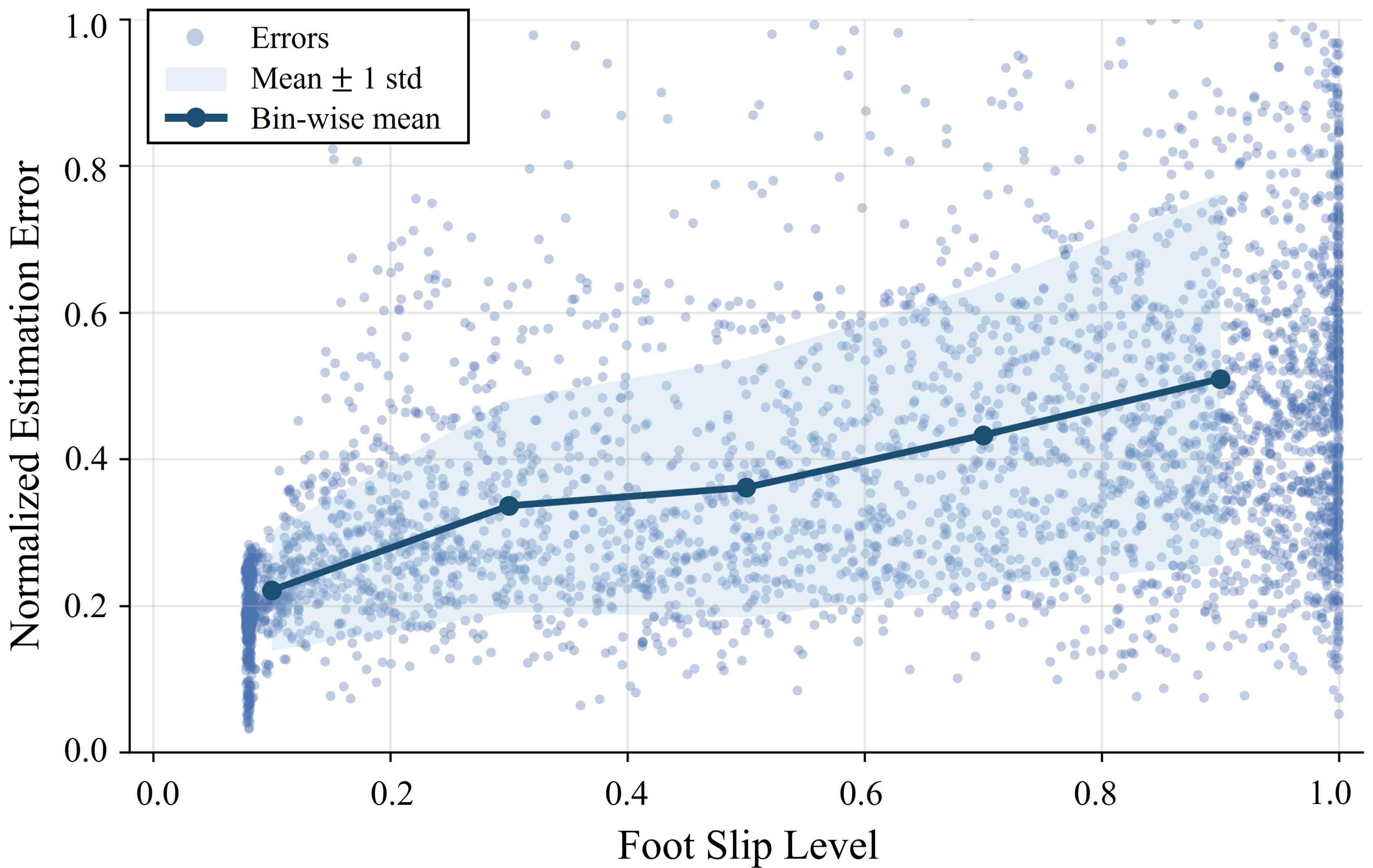}
    \caption{Relationship between foot slip level and normalized state estimation errors of the InEKF on the overall terrain. Dots denote per-timestep errors, and colored markers show the mean $\pm$ one standard deviation within slip.}
    \label{fig:correlation}
\end{figure}
From the related work of Bloesch \textit{et al}.~\cite{bloesch2013state}, it has been consistently reported that foot slip adversely affects state estimation in legged robots. To verify that the previously defined foot slip level function serves as a meaningful surrogate that reflects the relationship with state estimation error, we compute the actual InEKF state estimation error and plot its relationship with the foot slip level in Fig.~\ref{fig:correlation}. In this figure, the rotation error is given by $e_t^{\text{rot}} = \|\mathrm{Log}\!\left(\mathbf{R}^{GT}_t(\bar{\mathbf{R}}^{+}_t)^{\top}\right)^{\vee}\|_2$, while the velocity and position errors are defined as $e_t^{\text{vel}} = \|\mathbf{v}_t^{\text{GT}} - \bar{\mathbf{v}}_t^+\|_2$ and $e_t^{\text{pos}} = \|\mathbf{p}_t^{\text{GT}} - \bar{\mathbf{p}}_t^+\|_2$, respectively. Each error component is then normalized by its 95th-percentile value, and a scalar state error is defined as the square root of the sum of squares of the normalized position, velocity, and rotation errors. This scalar error is visualized as the mean value within each slip-level bin. Specifically, the foot slip level is discretized into bins of width 0.2, and for each bin, the mean and standard deviation of the state error are computed. The resulting plot shows that the normalized estimation error increases as the foot slip level grows.

Pearson correlation $r = 0.68$ ($r^2 \approx 0.46$) indicates a moderate-to-strong positive correlation. While approximately half of the error variance remains attributable to other factors (terrain geometry, sensor noise, modeling errors), the monotonic relationship supports using foot slip level as a meaningful context signal for the compensator.

\section{Attention-based Neural Compensator}
\label{section:attennkf}
In this section, we design the Attention-based Neural Compensator (AttenNC) that uses the continuous foot slip level defined in the previous section to make the InEKF compensation structure vary across slip regimes. Since the compensator behavior should change depending on the slip regime, we treat slip not as a simple additional input, but as a context that determines how much and in what way the InEKF history should be trusted. We first encode the InEKF posterior and slip sequence into task-specific latent representations using GRU-based autoencoders, which aim to mitigate sensitivity to trajectory-specific patterns and encourage the network to focus on representation before applying slip-conditioned modulation. We then choose attention as the mechanism that weights these latent features according to the slip context: the contribution of the latent features to the compensation applied to the residual is learned as a function of the slip regime. While the original InNKF entangles slip, state, and measurements into a single black-box mapping, the proposed AttenNC computes weights via cross-attention—using the slip latent as the query and the InEKF latent as the key/value—to obtain an explicit slip-conditioned compensation structure that selectively emphasizes parts of the InEKF latent under each slip regime.

\subsection{Training Process of Neural Compensator}
\begin{figure*}
    \centering
    \includegraphics[width=0.82\linewidth]{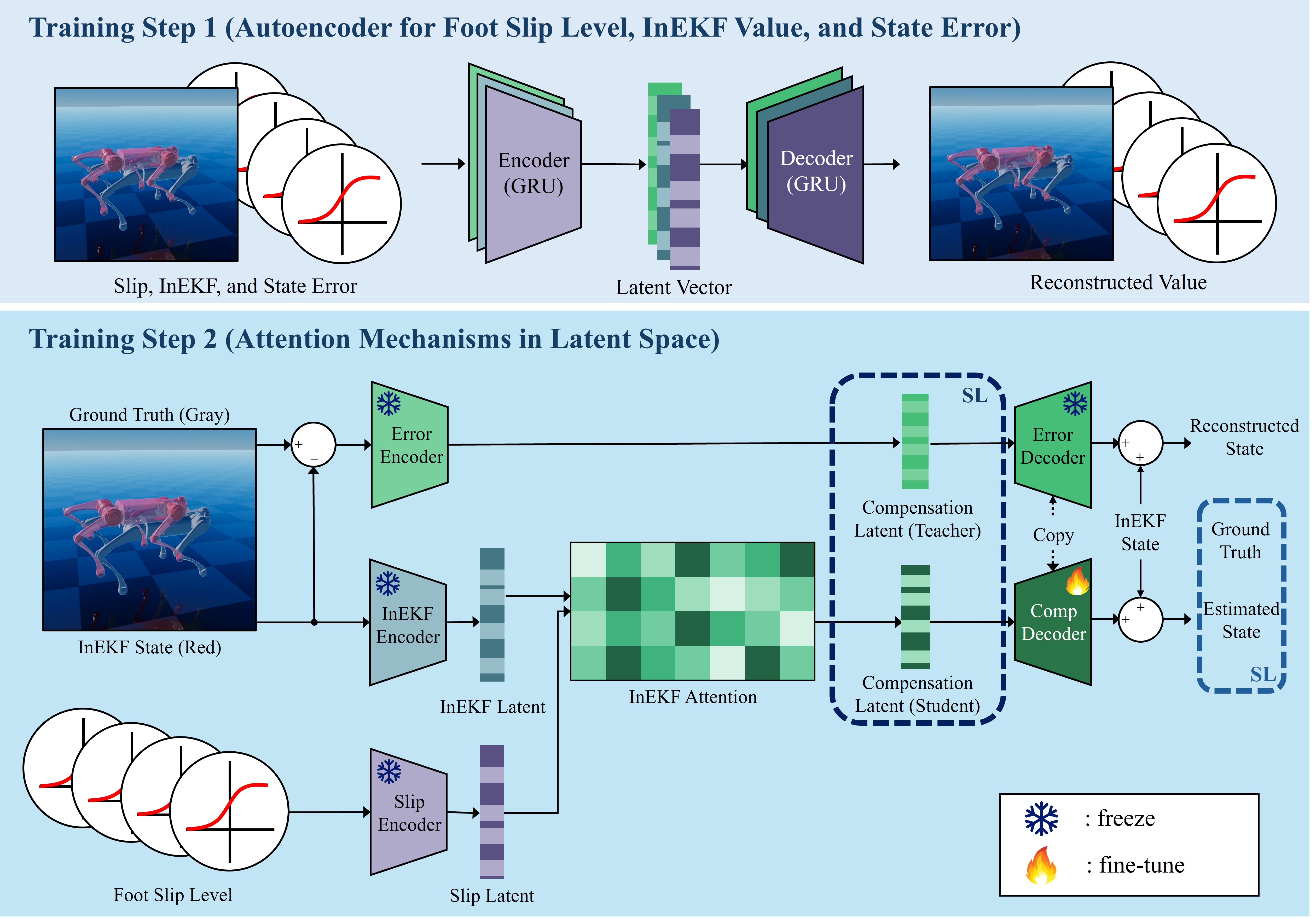}
    \caption{Training process of the Neural Compensator. It consists of two steps: Step 1 corresponds to autoencoder training, and Step 2 represents attention mechanism training. The snowflake symbol indicates a frozen model, while the fire symbol denotes a fine-tuned component.}
    \label{fig:training_process}
\end{figure*}

As illustrated in Fig.~\ref{fig:training_process}, the Attention-based NC (AttenNC) is trained as follows. The attention mechanism takes as inputs: (i) Foot Slip Level (the core signal in this work); (ii) InEKF-derived variables—(a) the InEKF update residual (update error), and (b) the normalized estimated state and its state error, where the state comprises $\textbf{R},\textbf{v}, \textbf{p}$. We include the estimated state as an additional context signal to guide the attention module by providing information about the robot's current motion regime, even though it is not expected to be strongly correlated with slip. Training Step 1 learns an autoencoder that maps these inputs to a latent space. Given sequences of length 50, we employ gated recurrent units (GRUs) \cite{cho2014learning} in both the encoder and decoder to better capture temporal structure. Autoencoder hyperparameters are summarized in Table~\ref{table:gru}.

In Training Step 2, the attention mechanism is trained in the latent space. The encoders pre-trained in Step 1 are utilized as frozen models during subsequent training. The Error Encoder generates a teacher latent vector, the InEKF Encoder and the slip Encoder each produce their own latent vectors. In the attention module, the slip latent is used as the query, whereas the InEKF latent supplies both keys and values.

\begin{table}[t]
\centering
\caption{GRU autoencoder hyperparameters by input feature}
\label{table:gru}
\begin{tabular}{|l|c|c|c|}
\hline
\textbf{Feature} & \textbf{Hidden dim.} & \textbf{Latent dim.} & \textbf{Layers} \\
\hline
Slip                    & 64  & 16 & 2 \\
InEKF Value             & 128 & 32 & 2 \\    
State Error             & 128 & 32 & 2 \\
\hline
\end{tabular}
\end{table}
The Error Encoder is used only during training and is excluded at inference. Therefore, it is not used in the attention module, but it serves as a teacher during training. Through the InEKF Attention mechanism, the compensation latent vector—used for actual state estimation—is generated as a student, and supervised learning (SL) is performed using the compensation latent vector generated by the Error Encoder as the teacher.

To further improve training efficiency, the estimated state obtained from the decoder is also trained via SL. Among the decoders trained in Step 1, the error decoder is fine-tuned and reused as the compensation decoder. The compensation latent vector is used as input to this decoder to predict the compensation error. This error is added to the InEKF-estimated state to yield the final estimated state.

An additional SL loss is applied using ground truth values available in the simulation. As a result, the total loss function used in Step 2 is defined as:
\[
\mathcal{L}_{\text{total}} = \lambda_1 \mathcal{L}_{\text{latent}} + \lambda_2 \mathcal{L}_{\text{state}},
\]
where $\mathcal{L}_{\text{latent}}$ represents the loss between the teacher and student latent vectors, and $\mathcal{L}_{\text{state}}$ denotes the error between the predicted and ground truth state. Each loss is computed using the following equations. We define the state error as
\begin{equation*}
\mathbf{e}_t \triangleq
\begin{bmatrix}
\mathrm{Log}\!\left(\mathbf{R}^{GT}_t(\bar{\mathbf{R}}^{+}_t)^{\top}\right)^{\vee} \\
\mathbf{v}^{GT}_t-\bar{\mathbf{v}}^{+}_t \\
\mathbf{p}^{GT}_t-\bar{\mathbf{p}}^{+}_t
\end{bmatrix}
\in \mathbb{R}^{9},
\end{equation*}
and use a weighted MSE:
\begin{equation*}
\mathcal{L}_{\mathrm{state}} =
w_R \left\|\mathbf{e}^R_t\right\|^2
+ w_v \left\|\mathbf{e}^v_t\right\|^2
+ w_p \left\|\mathbf{e}^p_t\right\|^2 .
\end{equation*}
where $\mathbf{e}^R_t$, $\mathbf{e}^v_t$, and $\mathbf{e}^p_t$ denote the rotation, velocity, and position components of $\mathbf{e}_t$, respectively. The latent loss $\mathcal{L}_{\text{latent}}$ is defined as
\begin{align*}
\mathcal{L}_{\text{latent}} =\; & \mathcal{L}_{\text{mse}}(\mathbf{z}, \hat{\mathbf{z}})
+ 0.1 \cdot \left(1 - \cos(\mathbf{z}, \hat{\mathbf{z}})\right) \\
& + 0.01 \cdot D_{\mathrm{KL}}\!\left(
\mathrm{softmax}(\hat{\mathbf{z}}) \,\|\, \mathrm{softmax}(\mathbf{z})
\right).
\end{align*}
Here, $\mathbf{z}$ and $\hat{\mathbf{z}}$ denote the teacher and student latent vectors, respectively. $\mathcal{L}_{\text{mse}}(\mathbf{z}, \hat{\mathbf{z}})$ is the MSE in the latent space, $\cos(\cdot, \cdot)$ represents the cosine similarity, $\mathcal{L}_{\text{mse}}$ is the mean square error loss, and $D_{\mathrm{KL}}(p \| q) = \sum p_i \log \frac{p_i}{q_i}$ denotes the Kullback-Leibler (KL) divergence between two distributions.

The training dataset is generated in Isaac Gym \cite{makoviychuk2isaac} using a Unitree Go1 \cite{unitree} model across randomized terrains and dynamics (friction, terrain shape, base mass, external pushes, and sensor noise). Each dataset episode lasts 200 s at 500 Hz.

\subsection{Attention-based Neural-augmented Kalman Filter}


This work proposes a hybrid state estimator that combines model-based and learning-based components, where the model-based part follows the InEKF structure proposed by Hartley \textit{et al}.~\cite{hartley2020contact} without modification. In the InEKF, the state is defined as $\mathbf{Z}_t \in \mathrm{SE}_{N+2}(3)$:
\begin{equation}
\mathbf{Z}_t \triangleq
\begin{bmatrix}
\mathbf{S}_{t} & {}^{\text{w}}\mathbf{d}_{\text{WC}_1}(t) & {}^{\text{w}}\mathbf{d}_{\text{WC}_2}(t) & \cdots & {}^{\text{w}}\mathbf{d}_{\text{WC}_N}(t)\\
\mathbf{0}_{1,5} & 1 & 0 & \cdots & 0 \\
\mathbf{0}_{1,5} & 0 & 1 & \cdots & 0 \\
\vdots  & \vdots  &  \vdots & \ddots & \vdots\\
\mathbf{0}_{1,5} & 0 & 0 & \cdots & 1\\
\end{bmatrix}
\label{eq:Z_t}
\end{equation}
where the base state $\mathbf{S}_t$ is given by
\[
\mathbf{S}_t =
\begin{bmatrix}
        \mathbf{R}_t & \mathbf{v}_t & \mathbf{p}_t \\
        \mathbf{0}_{1,3} & 1 & 0 \\
        \mathbf{0}_{1,3} & 0 & 1
\end{bmatrix}
\in \mathbb{R}^{5\times 5},
\]
with $\mathbf{R}_t \in \mathrm{SO}(3)$ denoting the rotation from the body frame to the world frame, and $\mathbf{v}_t$ and $\mathbf{p}_t$ denoting the base velocity and position in the world frame. The terms ${}^{\mathrm{w}}\mathbf{d}_{\mathrm{WC}_i}(t)$ represent the world-frame positions of each contact point, which are used as measurements in the InEKF. The prediction and update steps are all carried out on $\mathrm{SE}_{N+2}(3)$, and the detailed equations follow Hartley \textit{et al}.~\cite{hartley2020contact}.

From the posterior state $\bar{\mathbf{Z}}_t^+$ obtained by the InEKF, we first extract the rotation $\bar{\mathbf{R}}_t^+$, velocity $\bar{\mathbf{v}}_t^+$, and position $\bar{\mathbf{p}}_t^+$. We then represent the base orientation on the Lie algebra as $\bar{\boldsymbol{\theta}}_t^+ \triangleq \mathrm{Log}(\bar{\mathbf{R}}_t^+)^{\vee} \in \mathbb{R}^3$ where $(\cdot)^\vee$ denotes the vee operator and $\mathrm{Log}(\cdot)$ is the logarithmic map on $\mathrm{SO}(3)$. Accordingly, we define the 9-dimensional base state vector as
\begin{equation}
    \bar{\mathbf{x}}_t^+ \triangleq
    \begin{bmatrix}
        (\bar{\boldsymbol{\theta}}_t^+)^\top &  (\bar{\mathbf{v}}_t^+)^\top & (\bar{\mathbf{p}}_t^+)^\top
    \end{bmatrix}^\top \in \mathbb{R}^9.
\end{equation}
The NC applies a residual-style compensation to this state. Using local tangent-space $\delta$ representations, we parametrize the orientation compensation in the Lie algebra $\mathfrak{so}(3)$. Specifically, the NC predicts a small rotation residual $\delta\boldsymbol{\theta}_t\in \mathbb{R}^3$, which is applied to the base orientation on $\mathrm{SO}(3)$ via the exponential map (left action). The velocity and position components are compensated additively in $\mathbb{R}^3$. 

The AttenNC takes $\bar{\mathbf{x}}_t^+$ as input and predicts a residual compensation $\bar{\mathbf{e}}_t \in \mathbb{R}^9$, which we decompose as $\bar{\textbf{e}}_t\triangleq [\delta\bar{\boldsymbol{\theta}}_t^\top, \delta\bar{\textbf{v}}_t^\top, \delta\bar{\textbf{p}}_t^\top]^\top$. We then apply the correction as
\begin{equation}
\bar{\mathbf{R}}_{t}^{++} = \mathrm{Exp}\!\left(\delta\boldsymbol{\theta}_{t}^{\wedge}\right)\bar{\mathbf{R}}_{t}^{+}, \quad
\begin{bmatrix}\bar{\mathbf{v}}_{t}^{++}\\ \bar{\mathbf{p}}_{t}^{++}\end{bmatrix}
=
\begin{bmatrix}\bar{\mathbf{v}}_{t}^{+}\\ \bar{\mathbf{p}}_{t}^{+}\end{bmatrix}
+
\begin{bmatrix}\delta\bar{\mathbf{v}}_{t}\\ \delta\bar{\mathbf{p}}_{t}\end{bmatrix}.
\label{eq:final_correction}
\end{equation}
where $(\cdot)^{\wedge}$ denotes the hat operator and $\mathrm{Exp}(\cdot)$ is the exponential map on $\mathrm{SO}(3)$. For reporting in $\mathbb{R}^9$, we set $\bar{\textbf{x}}_t^{++}\triangleq[\mathrm{Log}(\bar{\textbf{R}}_t^{++})^{\vee^\top}, \bar{\textbf{v}}_t^{{++}^\top}, \bar{\textbf{p}}_t^{{++}^\top}]^\top$. The NC is a mean-only compensation module that adds a residual to the base state extracted from the InEKF posterior, without modifying the covariance matrix computed by the InEKF. The underlying InEKF performs prediction, update on $\mathrm{SE}_{N+2}(3)$ exactly as in the original formulation, and the NC output is not fed back into the InEKF recursion. In other words, the AttenNC post-processes the InEKF posterior $\bar{\mathbf{x}}_t^+$ with a slip-aware compensation to produce $\bar{\mathbf{x}}_t^{++}$ as the final reported estimation, while we retain the InEKF covariance (not adjusted to the corrected mean $\bar{\mathbf{x}}_t^{++}$).

\begin{figure}
    \centering
    \includegraphics[width=0.75\linewidth]{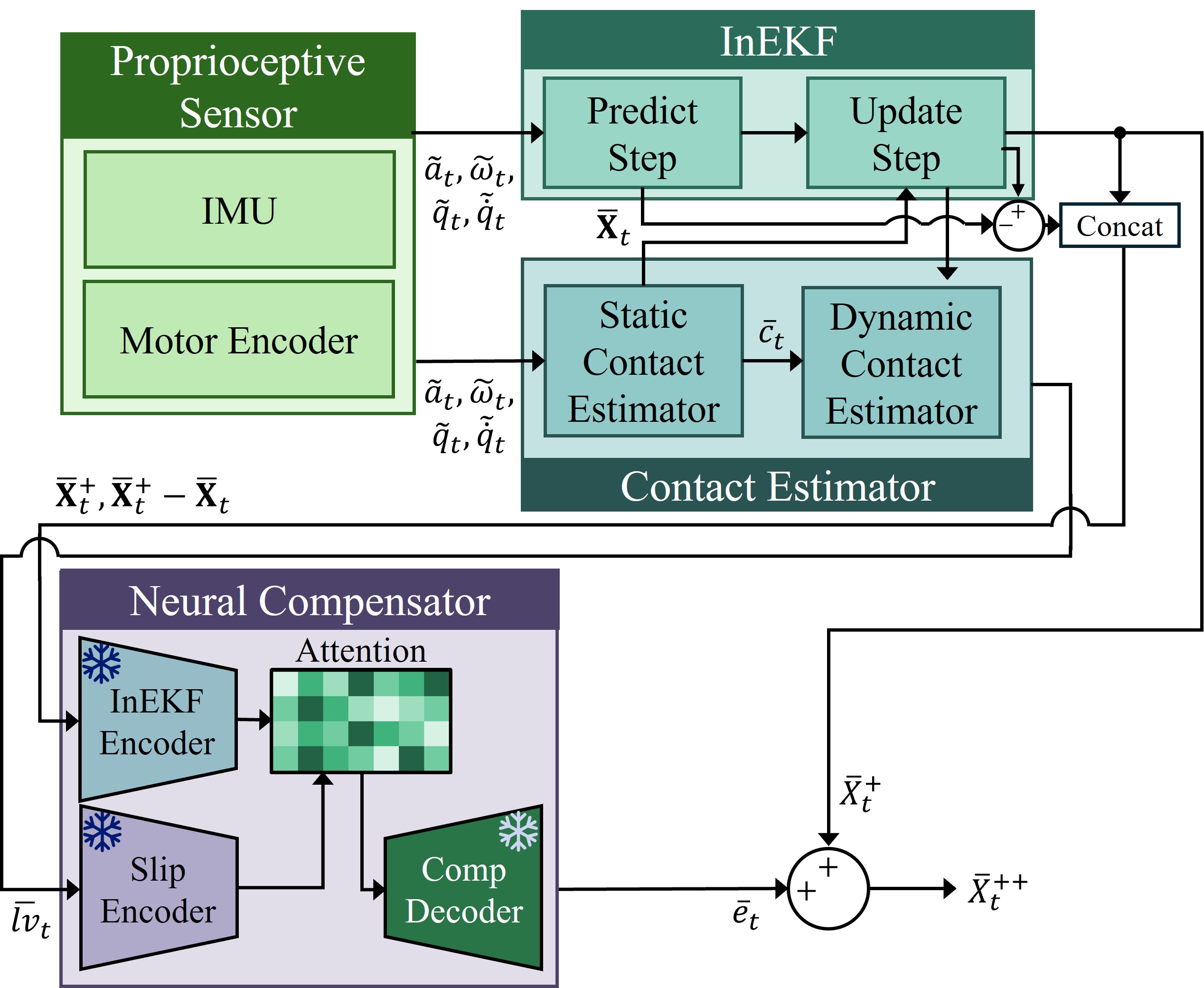}
    \caption{Structure of the AttenNKF. The Neural Compensator, composed of the Encoder-Decoder and Attention modules trained in Section~\ref{section:attennkf}-A, is augmented into the InEKF to generate the compensated state.}
    \label{fig:attennkf}
\end{figure}
The compensation term $\bar{\mathbf{e}}_t$ is obtained through the AttenNC. The AttenNC encodes the InEKF history and slip history with separate GRU encoders and applies a scaled dot-product cross-attention so that the InEKF latent is modulated by the slip condition. The InEKF encoder $E_\varphi(\cdot)$ takes as input the history of the InEKF base state $\bar{\mathbf{x}}_{t-H:t}^+$ and the internal InEKF correction $\Delta \bar{\mathbf{x}}_{t-H:t}$ where $\Delta\bar{\textbf{x}}_\tau\triangleq[\mathrm{Log}(\bar{\textbf{R}}_\tau^+\bar{\textbf{R}}_\tau^{-1})^{\vee^\top}, (\bar{\textbf{v}}_\tau^+-\bar{\textbf{v}}_\tau)^\top, (\bar{\textbf{p}}_\tau^+-\bar{\textbf{p}}_\tau)^\top]^\top$:
\begin{equation}
\mathbf{Z}_t^{\text{InEKF}}
= E_\varphi\big(\bar{\mathbf{x}}_{t-H:t}^+, \Delta\bar{\mathbf{x}}_{t-H:t}\big)
\in \mathbb{R}^{H \times d_{\text{InEKF}}},
\end{equation}
and produces an InEKF latent sequence, where $H$ is the history length and $d_{\text{InEKF}}$ is the InEKF latent dimension. Similarly, the slip encoder $E_\rho(\cdot)$ takes the continuous slip-level history $\bar{lv}_{t-H:t}$ (4-dimensional slip level over four feet) and generates
\begin{equation}
\mathbf{Z}_t^{\text{slip}}
= E_\rho\big(\bar{lv}_{t-H:t}\big)
\in \mathbb{R}^{H \times d_{\text{slip}}},
\end{equation}
which is the slip latent sequence.

In the cross-attention module, the slip latent acts as the query and the InEKF latent as the key/value. Over the history window, we define
\begin{equation}
\begin{aligned}
\mathbf{Q}_t &= \mathbf{Z}_t^{\text{slip}} \mathbf{W}_q \in \mathbb{R}^{H \times d_h},\\
\mathbf{K}_t &= \mathbf{Z}_t^{\text{InEKF}} \mathbf{W}_k \in \mathbb{R}^{H \times d_h},\\
\mathbf{V}_t &= \mathbf{Z}_t^{\text{InEKF}} \mathbf{W}_v \in \mathbb{R}^{H \times d_h},
\end{aligned}
\end{equation}
where $\mathbf{W}_q \in \mathbb{R}^{d_{\text{slip}}\times d_h}$ and $\mathbf{W}_k,\mathbf{W}_v \in \mathbb{R}^{d_{\text{InEKF}}\times d_h}$ are learnable weights, and $d_h$ is the attention hidden dimension. The scaled dot-product attention is given by
\begin{equation}
\boldsymbol{\alpha}_t
= \mathrm{softmax}\left(\frac{1}{\sqrt{d_h}} \mathbf{Q}_t \mathbf{K}_t^\top\right)
\in \mathbb{R}^{H \times H},
\end{equation}
softmax is applied along the key dimension, producing per-query weights over the history window. The slip-conditioned InEKF latent sequence is obtained as
\begin{equation}
\mathbf{H}_t^{\text{att}}
= \boldsymbol{\alpha}_t \mathbf{V}_t
\in \mathbb{R}^{H \times d_h}.
\end{equation}

This context sequence is passed through a small MLP $f_{\text{MLP}}(\cdot)$ at each time step,
\begin{equation}
\mathbf{Z}_t^{\text{comp}}
= f_{\text{MLP}}\big(\mathbf{H}_t^{\text{att}}\big)
\in \mathbb{R}^{H \times d_z},
\end{equation}
and then into a GRU-based decoder $D_\nu(\cdot)$ to predict the compensation:
\begin{equation}
\bar{\mathbf{e}}_t
= D_\nu\big(\mathbf{Z}_t^{\text{comp}}\big)
\in \mathbb{R}^9.
\end{equation}
The decoder produces a sequence of compensations, and the last time-step output is used as the compensation at time $t$ in (\ref{eq:final_correction}), yielding a slip-conditioned compensation term that adapts across different slip regimes.

\section{EXPERIMENTS}
\label{section:experiments}
The AttenNKF proposed in this paper aims to improve state estimation performance by focusing on foot slip, a key factor in legged robots. Therefore, experiments were conducted under conditions where foot slip frequently occurs. For comparison, we evaluate state estimation performance against slip-focused baselines: Slip Rejection (SR) \cite{kim2021legged}, the Learned Contact (LC) event–based method \cite{lin2022legged}, and the Invariant Neural-Augmented Kalman Filter (InNKF) with a Neural Compensator (NC) \cite{lee2025legged}. The experiments were conducted in both indoor and outdoor environments using a learning-based blind locomotion controller \cite{rudin2022learning}. Before conducting real-world evaluation, to reduce the sim-to-real gap, all learning-based models (LC, InNKF, AttenNKF) are fine-tuned on the same real subset of walking data collected on only flat and gravel field terrains in the real world. The real-world fine-tuning uses sequences disjoint from all evaluation runs. In Table~\ref{table:comparison_results}--\ref{table:ablation1}, we report relative errors such as rotation, velocity, and position ($\textbf{RE}_\text{rot}, \textbf{RE}_\text{vel}, \textbf{RE}_\text{pos}$) defined over a 5\,m traveled distance, where each entry is the mean RE on that terrain with the standard deviation in brackets, and $\textbf{RE}_\text{rot}$, $\textbf{RE}_\text{vel}$, and $\textbf{RE}_\text{pos}$ are expressed in deg, m/s, and m, respectively.

\subsection{Indoor Evaluation under Slip Conditions}
\begin{figure}
    \centering
    \includegraphics[width=0.85\linewidth]{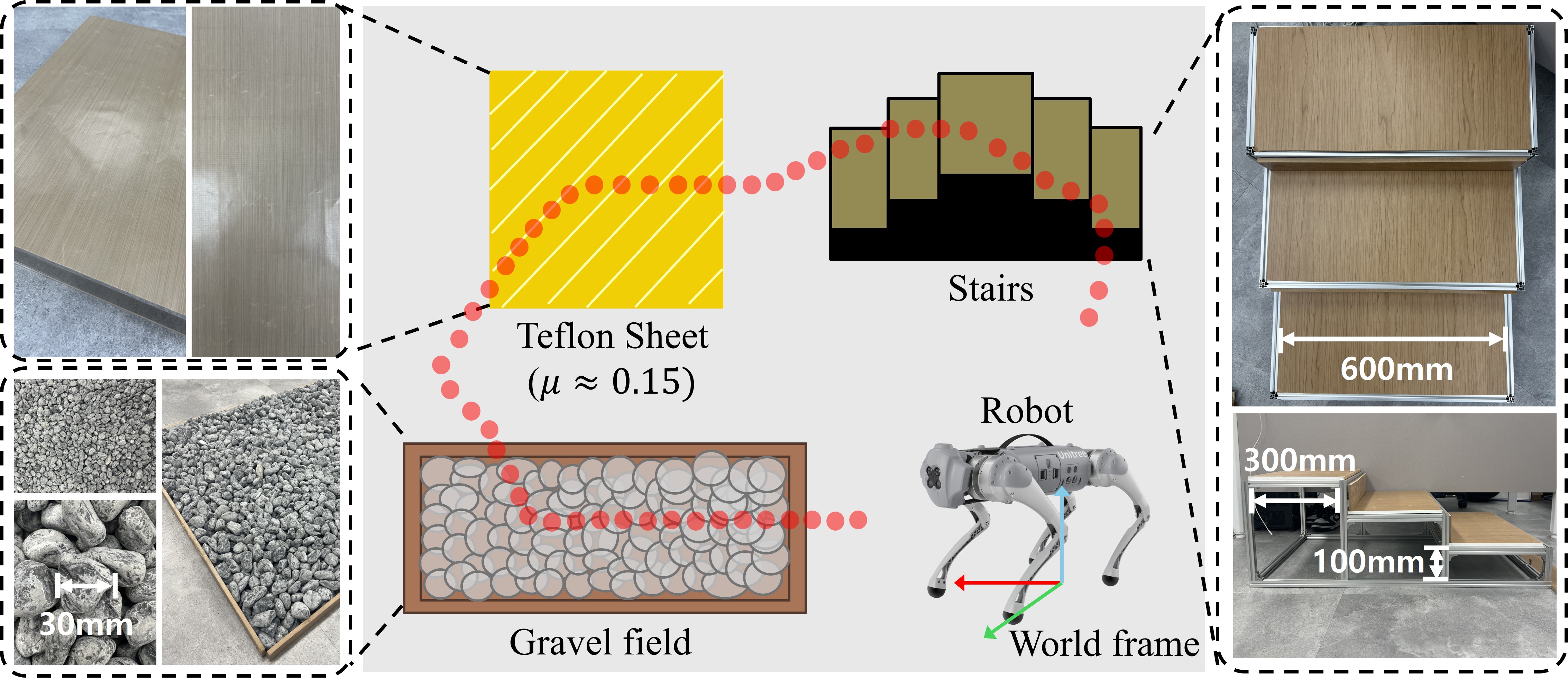}
    \caption{Real-world scenario of the indoor experimental environment, consisting of: (1) a gravel field with 30 mm-diameter pebbles, (2) a Teflon sheet with a low friction coefficient (0.10–0.15), and (3) a three-tier staircase (60 cm width, 30 cm depth, 10 cm height per step) designed for both uphill and downhill traversal.}
    \label{fig:indoor}
\end{figure}

\begin{table}[t!]
\centering
{
\caption{Relative rotation (deg), velocity (m/s), and position (m) errors over 5 m across various terrains; Mean (std)}
\resizebox{\columnwidth}{!}{%
\begin{tabular}{c|c|c|c|c|c|c}
\hline
\multicolumn{2}{c|}{\textbf{Terrain}} & \textbf{Metric} & \textbf{SR} \cite{kim2021legged} & \textbf{LC} \cite{lin2022legged} & \textbf{InNKF} \cite{lee2025legged} & \textbf{AttenNKF} \\
\hline
\multirow{30}{*}{ID} 
& \multirow{6}{*}{Flat}
& \multirow{2}{*}{\textbf{RE}$_{\text{rot}}$} & 1.542 & 1.589 & 2.774 & \textbf{1.254} \\
& &                                   & (0.650) & (0.692) & (1.589) & (0.966) \\
& & \multirow{2}{*}{\textbf{RE}$_{\text{vel}}$} & 0.538 & 0.428 & 0.715 & \textbf{0.417} \\
& &                                   & (0.039) & (0.029) & (0.332) & (0.072) \\
& & \multirow{2}{*}{\textbf{RE}$_{\text{pos}}$} & 0.388 & 0.321 & 0.389 & \textbf{0.214} \\
& &                                   & (0.034) & (0.027) & (0.194) & (0.084) \\
\cline{2-7}

& \multirow{6}{*}{\shortstack{Gravel\\Field}}
& \multirow{2}{*}{\textbf{RE}$_{\text{rot}}$} & 1.667 & 1.712 & 2.588 & \textbf{1.150} \\
& &                                   & (0.954) & (0.954) & (1.125) & (0.474) \\
& & \multirow{2}{*}{\textbf{RE}$_{\text{vel}}$} & 0.302 & 0.428 & 0.650 & \textbf{0.165} \\
& &                                   & (0.192) & (0.370) & (0.408) & (0.120) \\
& & \multirow{2}{*}{\textbf{RE}$_{\text{pos}}$} & 0.700 & 1.291 & 0.706 & \textbf{0.048} \\
& &                                   & (0.412) & (1.034) & (0.510) & (0.043) \\
\cline{2-7}

& \multirow{6}{*}{\shortstack{Teflon\\Sheet}}
& \multirow{2}{*}{\textbf{RE}$_{\text{rot}}$} & 1.660 & 1.694 & 1.958 & \textbf{1.352} \\
& &                                   & (1.152) & (1.158) & (0.934) & (0.734) \\
& & \multirow{2}{*}{\textbf{RE}$_{\text{vel}}$} & 0.483 & \textbf{0.422} & 0.527 & 0.423 \\
& &                                   & (0.272) & (0.280) & (0.368) & (0.121) \\
& & \multirow{2}{*}{\textbf{RE}$_{\text{pos}}$} & 0.610 & 0.506 & 0.872 & \textbf{0.260} \\
& &                                   & (0.118) & (0.224) & (0.351) & (0.140) \\
\cline{2-7}

& \multirow{6}{*}{Stairs}
& \multirow{2}{*}{\textbf{RE}$_{\text{rot}}$} & 3.452 & 3.494 & 7.812 & \textbf{3.346} \\
& &                                   & (0.859) & (0.868) & (3.325) & (0.902) \\
& & \multirow{2}{*}{\textbf{RE}$_{\text{vel}}$} & 0.568 & 0.452 & 1.878 & \textbf{0.174} \\
& &                                   & (0.027) & (0.027) & (0.086) & (0.017) \\
& & \multirow{2}{*}{\textbf{RE}$_{\text{pos}}$} & 0.348 & 0.414 & 0.086 & \textbf{0.081} \\
& &                                   & (0.016) & (0.010) & (0.029) & (0.016) \\
\cline{2-7}

& \multirow{6}{*}{\shortstack{Overall\\Scenario}}
& \multirow{2}{*}{\textbf{RE}$_{\text{rot}}$} & 2.916 & 2.992 & 4.106 & \textbf{2.143} \\
& &                                   & (1.209) & (1.217) & (2.470) & (1.357) \\
& & \multirow{2}{*}{\textbf{RE}$_{\text{vel}}$} & 0.732 & 0.684 & 0.770 & \textbf{0.683} \\
& &                                   & (0.462) & (0.425) & (0.316) & (0.373) \\
& & \multirow{2}{*}{\textbf{RE}$_{\text{pos}}$} & 0.455 & 0.554 & 0.471 & \textbf{0.051} \\
& &                                   & (0.263) & (0.313) & (0.323) & (0.039) \\
\hline

\multirow{6}{*}{OOD}
& \multirow{6}{*}{\shortstack{Soft\\Terrain}}
& \multirow{2}{*}{\textbf{RE}$_{\text{rot}}$} & 4.167 & 4.243 & 4.582 & \textbf{1.478} \\
& &                                   & (2.369) & (2.371) & (1.920) & (0.342) \\
& & \multirow{2}{*}{\textbf{RE}$_{\text{vel}}$} & 0.622 & 0.565 & 0.665 & \textbf{0.325} \\
& &                                   & (0.306) & (0.372) & (0.393) & (0.153) \\
& & \multirow{2}{*}{\textbf{RE}$_{\text{pos}}$} & 0.539 & 0.979 & 0.545 & \textbf{0.314} \\
& &                                   & (0.217) & (0.667) & (0.212) & (0.123) \\
\hline

\end{tabular}
}
\label{table:comparison_results}
}
\end{table}

\begin{figure*}
    \centering
    \includegraphics[width=0.9\linewidth]{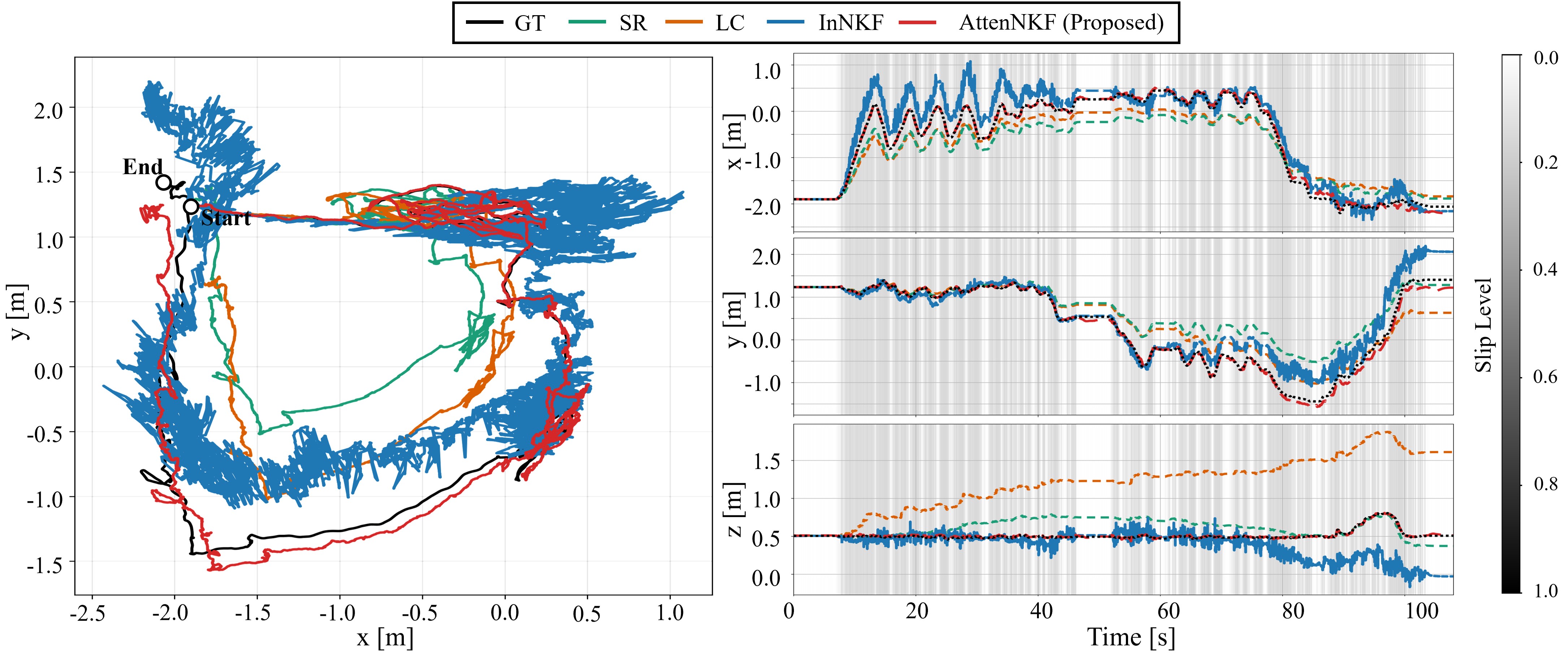}
    \caption{The state estimation results of the indoor experiments. Left: trajectories on the $x$–$y$ plane. Right: time histories of the estimated states ($x$, $y$, $z$); background shading denotes foot slip level (0–1). Color legend: black—Ground Truth (GT); green—Slip Rejection (SR); orange—Learned Contact (LC); blue—Invariant Neural-Augmented Kalman Filter (InNKF); red—Attention-Based Neural-Augmented Kalman Filter (AttenNKF, proposed method).}
    \label{fig:indoor_exp}
\end{figure*}

As illustrated in Fig.~\ref{fig:indoor}, we configure an indoor testbed comprising (i) a flat terrain, (ii) a gravel field made of 30\,mm pebbles to induce frequent slip, (iii) a flat Teflon sheet representing a highly slippery surface with friction coefficient $\mu \approx 0.15$, and (iv) stairs to emulate blind locomotion where unexpected footstep placement can occur. Ground truth is recorded with a Vicon motion-capture system at 100\,Hz. The walking protocol is as follows: on the gravel field, the robot traverses $\sim$1.5\,m back-and-forth five times; on the Teflon sheet, three round trips are performed; and on the stairs, one ascent and one descent are executed.

Fig.~\ref{fig:indoor_exp} compares the proposed method with the three baselines across all four terrains in terms of position estimation. Without any loop-closure or additional post-processing, AttenNKF shows smaller deviation from the ground truth in Fig.~\ref{fig:indoor_exp}. In particular, the right-hand plots show that, unlike SR, LC, and InNKF—whose errors grow with increasing foot slip level—AttenNKF shows smaller drift under high slip. Although InNKF employs an NC, its behavior can be brittle in situations not represented in the data used for training, occasionally exhibiting large spikes (see Fig.~\ref{fig:indoor_exp}). By contrast, the proposed method exhibits fewer such outliers.

As shown in Table~\ref{table:comparison_results}, on the flat terrain where slip is limited, all methods exhibit similar performance, since AttenNC outputs compensations that remain close to zero for most of the trajectory in such low-slip regimes, causing AttenNKF to behave almost identically to the baselines. In contrast, on the gravel field where slip occurs frequently, the difference becomes pronounced and AttenNKF reduces $\textbf{RE}_\text{pos}$ by $93.1\%$ relative to the strongest baseline. On the low-friction Teflon sheet, slip is much more significant and all methods exhibit larger errors, but AttenNKF still yields the lowest position RE, reducing $\textbf{RE}_\text{pos}$ by $48.6\%$ compared to the best-performing baseline. On the stairs, where the contact configuration continuously changes, SR and LC show large drift, while AttenNKF attains smaller RE in all components; in particular, $\textbf{RE}_\text{pos}$ is reduced by $5.8\%$ compared to InNKF, whereas rotational RE remains at a similar level for SR, LC, and AttenNKF.

Finally, to examine behavior on an out-of-distribution (OOD) terrain that is not used for training, we evaluate on a soft terrain excluded from the training data. On this terrain, LC and InNKF show degraded performance, whereas AttenNKF achieves smaller errors for all three metrics; in particular, $\textbf{RE}_\text{pos}$ decreases by $41.7\%$ relative to SR, suggesting that the slip-conditioned structure partially transfers to terrains not used during training.

\subsection{Outdoor Experiment}
\begin{figure}
    \centering
    \includegraphics[width=0.83\linewidth]{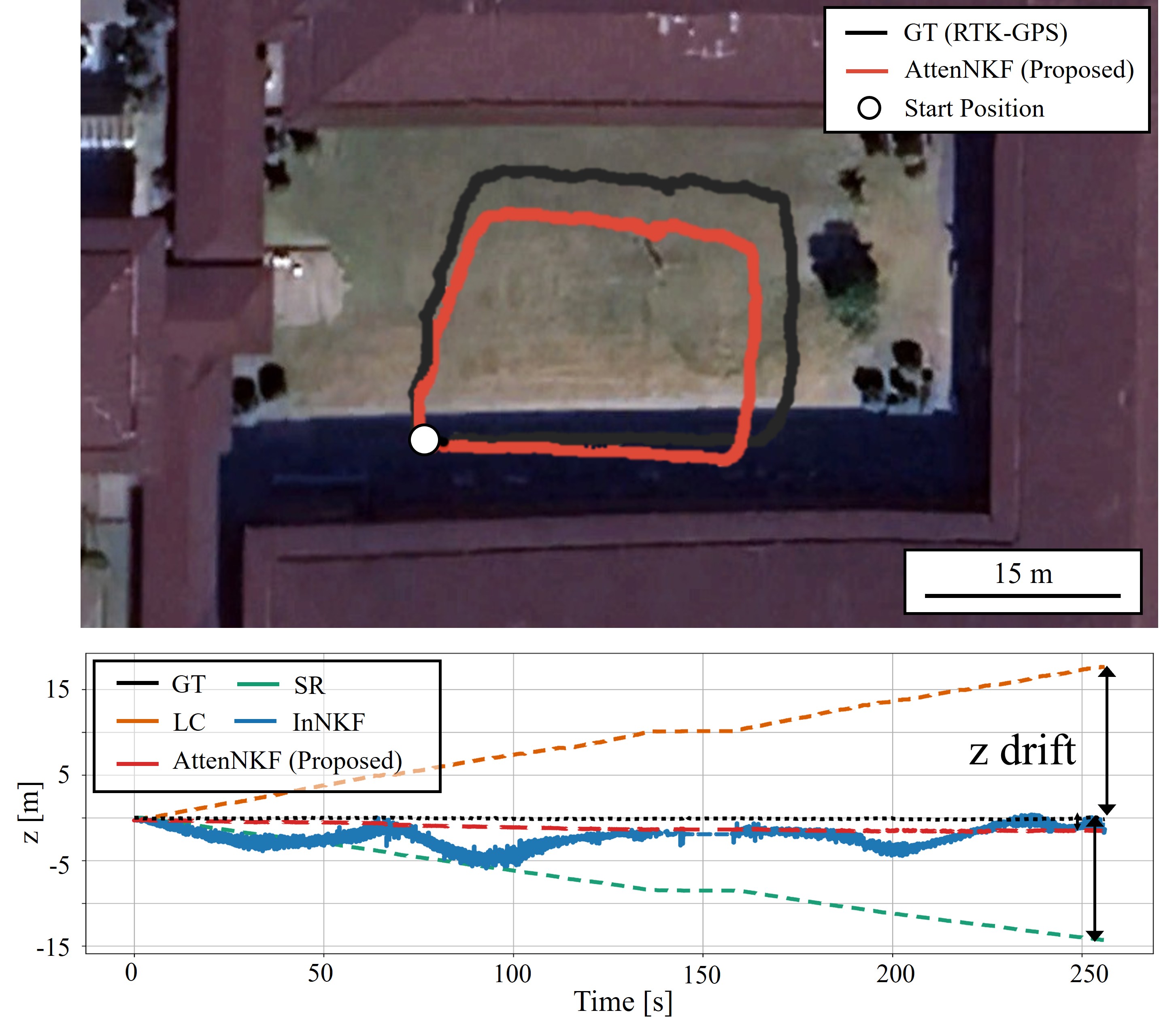}
    \caption{Outdoor experiment results. Top: Google Earth view of the trajectory. Bottom: estimated vertical position ($z$) over time.}
    \label{fig:outdoor_result}
\end{figure}
\begin{table}[t!]
\centering
\caption{Relative Position Errors (over a 5 m distance) across Grass; Mean (std)}
\resizebox{\columnwidth}{!}{%
\begin{tabular}{c|c|c|c|c|c}
\hline
\textbf{Terrain} & \textbf{Metric} & \textbf{SR \cite{kim2021legged}} & \textbf{LC \cite{lin2022legged}} & \textbf{InNKF \cite{lee2025legged}} & \textbf{AttenNKF} \\
\hline
\multirow{2}{*}{Outdoor}
& \multirow{2}{*}{\textbf{RE}$_{\text{pos}}$} & 0.347 & 0.409 & 0.582 & \textbf{0.155} \\
&                                   & (0.019) & (0.026) & (0.311) & (0.040) \\
\hline
\end{tabular}
}
\label{table:outdoor}
\end{table}
To verify that the proposed AttenNKF remains effective not only on indoor trajectories of a few meters but also over longer distances and irregular terrain, we conducted an outdoor walking experiment of approximately 5 minutes on a deformed grass field covering about 100\,m. We employed a Real-Time Kinematic Global Positioning System (RTK-GPS) as the ground truth. The position relative error $\mathbf{RE}_\text{pos}$, evaluated together with the three baseline methods used in Section~\ref{section:experiments}-A, is summarized in Table~\ref{table:outdoor}. In the outdoor segment, AttenNKF reduces $\mathbf{RE}_\text{pos}$ by about 55\%, 62\%, and 73\% of those of SR, LC, and InNKF, respectively, and achieves the smallest error in both mean and variance. In the top plot of Fig.~\ref{fig:outdoor_result}, the forward distance of AttenNKF is slightly underestimated compared to the ground truth, resulting in a residual $x$–$y$ position error at the loop. This can be interpreted as a systematic negative bias in the forward velocity estimate caused by persistent foot sinking and slipping on the soft ground, which violates the zero-velocity contact assumption; AttenNC effectively suppresses slip-induced instantaneous drift, but does not completely remove the long-range forward distance scaling error that accumulates over the entire trajectory. On the other hand, as shown in the time histories at the bottom of Fig.~\ref{fig:outdoor_result}, SR, LC, and InNKF accumulate up to 10\,m of vertical ($z$-axis) drift over about 250\,s, whereas the proposed method remains within approximately $\pm 2$–3\,m around the reference trajectory, demonstrating that vertical drift is substantially mitigated even during extended outdoor walking.

\subsection{Ablation Study}
Ablation is evaluated on the same 25\,m indoor sequence (overall scenario) used in Section~\ref{section:experiments}-A. This sequence contains low-slip (flat), medium-slip (gravel), high-slip (Teflon), and stair segments with frequently changing contact configurations, so that various slip regimes can be assessed within a single dataset. To isolate the effect of the architecture, we compare three variants of the proposed NC—without the attention module \textit{(No-Atten)}, without slip conditioning \textit{(Self-Atten)}, and without the teacher latent supervision \textit{(No-Teach)}—all trained on the same data, with the same loss functions and hyperparameters, differing only in network structure.

\begin{table}[t!]
\centering
\caption{Ablation Study: Effect of the Teacher-Student and Cross-Attention Architecture of the AttenNC; Mean (std)}
\resizebox{\columnwidth}{!}{%
\begin{tabular}{c|c|c|c|c|c}
\hline
\textbf{Terrain} & \textbf{Metric} & \textbf{No-Atten} & \textbf{Self-Atten} & \textbf{No-Teach} & \textbf{Proposed} \\
\hline
\multirow{6}{*}{\shortstack{Overall\\Scenario}}
& \multirow{2}{*}{\textbf{RE}$_{\text{rot}}$} & 2.791 & 2.644 & 3.898 & \textbf{2.143} \\
&                                    & (1.348) & (1.286) & (1.839) & (1.357) \\
& \multirow{2}{*}{\textbf{RE}$_{\text{vel}}$} & 0.809 & 0.878 & 0.842 & \textbf{0.683} \\
&                                    & (0.534) & (0.495) & (0.603) & (0.373) \\
& \multirow{2}{*}{\textbf{RE}$_{\text{pos}}$} & 0.208 & 0.266 & 0.263 & \textbf{0.051} \\
&                                   & (0.174) & (0.195) & (0.128) & (0.039) \\
\hline
\end{tabular}
}
\label{table:ablation1}
\end{table}
First, to assess the impact of the attention mechanism, we examine the No-Atten variant of the NC. This variant concatenates the slip and InEKF histories and directly regresses the error. As shown in Table~\ref{table:ablation1}, some compensation is achieved even without attention, but the position RE remains much larger than that of the proposed method on the same 25 m indoor sequence (0.208 vs 0.051). Under identical data and training settings, removing cross-attention therefore leads to a larger position error on this run, suggesting that reweighting the InEKF representation according to the slip condition is beneficial for reducing drift.

In contrast, the Self-Atten NC uses self-attention only on the InEKF latent without slip, and its $\textbf{RE}_\text{pos}$ is also significantly larger than that of AttenNKF (0.266 vs 0.051); this is consistent with the intuition that, without explicit slip context, it is harder to distinguish between different slip regimes and error patterns. In addition, in the variant without teacher guidance for the latent, the NC exhibits the largest rotation RE among the ablations, suggesting that the teacher signal helps the latent space capture the error structure more reliably.

\section{CONCLUSIONS}
\label{section:conclusion}
In this study, we proposed the AttenNKF, which augments the InEKF with a NC designed based on an attention mechanism. By using foot slip as the attention query, the proposed method introduces an explicit slip-conditioned modulation for the neural compensator and improves performance relative to InNKF on slip-prone terrains. Across the indoor and outdoor evaluations, AttenNKF reduces Relative Error (RE) compared to SR, LC, and InNKF, and it maintains competitive accuracy on a $\sim$100\,m outdoor trajectory over deformable grass terrain. Furthermore, the proposed estimator operates at 580 Hz, confirming its suitability for real-time applications.

In future work, we plan to apply the AttenNKF-estimated states to autonomous control of legged robots. Moreover, we aim to extend the framework to other mobile robotic systems, such as aerial robots and wheeled robots, by identifying and learning the primary error-inducing factors in those platforms. This will allow us to evaluate the general applicability of AttenNKF across various robotic systems. In addition, we plan to extend the framework so that the NC not only corrects the mean state but also adjusts or estimates the associated covariance, aiming for the reported uncertainty to remain consistent with the corrected state. Finally, we will investigate tightly coupled formulations that feed the learned correction back into the filter recursion while preserving the underlying Lie-group structure.





\section*{Appendix}

\begin{table}[t!]
\centering
\caption{Ablation Study II: End-to-end training vs. the proposed two-stage training pipeline for AttenNC}
\resizebox{\columnwidth}{!}{%
\begin{tabular}{c|c|c|c}
\hline
\textbf{\makecell{Terrain}} &
\textbf{\makecell{Metric}} &
\textbf{End-to-End} &
\textbf{Proposed Pipeline} \\
\hline
\multirow{6}{*}{\shortstack{Overall\\Scenario}}
& \multirow{2}{*}{\textbf{RE}$_{\text{rot}}$} & 3.574 & \textbf{2.143} \\
&                                    & (1.358) & (1.357) \\
& \multirow{2}{*}{\textbf{RE}$_{\text{vel}}$} & 1.393 & \textbf{0.683} \\
&                                    & (0.899) & (0.373) \\
& \multirow{2}{*}{\textbf{RE}$_{\text{pos}}$} & 0.320 & \textbf{0.051} \\
&                                    & (0.135) & (0.039) \\
\hline
\end{tabular}
}
\label{table:ablation2}
\end{table}

\begin{table}[t!]
\centering
\caption{Ablation Study III: Comparison of slip-related context inputs (raw foot velocity, explicit slip level, and implicit slip encoding)}
\resizebox{\columnwidth}{!}{%
\begin{tabular}{c|c|c|c|c}
\hline
\textbf{\makecell{Terrain}} &
\textbf{\makecell{Metric}} &
\textbf{\makecell{Raw Foot\\Velocity}} &
\textbf{\makecell{Explicit\\Slip}} &
\textbf{\makecell{Implicit Slip\\(Proposed)}} \\
\hline
\multirow{6}{*}{\shortstack{Soft\\Terrain}}
& \multirow{2}{*}{\textbf{RE}$_{\text{rot}}$} & 3.759 & 4.241 & \textbf{1.478} \\
&                                    & (2.940) & (2.426) & (0.342) \\
& \multirow{2}{*}{\textbf{RE}$_{\text{vel}}$} & 1.736 & 0.588 & \textbf{0.325} \\
&                                    & (0.654) & (0.180) & (0.153) \\
& \multirow{2}{*}{\textbf{RE}$_{\text{pos}}$} & 1.056 & 0.908 & \textbf{0.314} \\
&                                    & (0.567) & (0.703) & (0.123) \\
\hline
\end{tabular}
}
\label{table:ablation3}
\end{table}

To verify the effectiveness of the proposed training framework, we compare end-to-end training against the proposed pipeline in Table~\ref{table:ablation2}. Since the end-to-end approach jointly learns both the mapping to the latent space and the attention module for state-error estimation, the optimization becomes more challenging, which can hinder training stability and overall performance. As shown in Table~\ref{table:ablation2}, the proposed pipeline consistently improves all metrics, reducing \textbf{RE}$_{\text{rot}}$ from 3.574 to 2.143 (40.0\%), \textbf{RE}$_{\text{vel}}$ from 1.393 to 0.683 (51.0\%), and \textbf{RE}$_{\text{pos}}$ from 0.320 to 0.051 (84.1\%).

We further conduct an ablation study to evaluate alternative slip-related context inputs, as summarized in Table~\ref{table:ablation3}. Since the implicit slip encoding is introduced to mitigate overfitting to raw foot-velocity signals or handcrafted slip levels, we evaluate the variants on an out-of-distribution setting (soft terrain). As shown in Table~\ref{table:ablation3}, the proposed implicit slip achieves the best results across all metrics. In particular, implicit slip reduces \textbf{RE}$_{\text{rot}}$ from 3.759 to 1.478 (60.7\%), \textbf{RE}$_{\text{vel}}$ from 1.736 to 0.325 (81.3\%), and \textbf{RE}$_{\text{pos}}$ from 1.056 to 0.314 (70.3\%) compared to using raw foot velocity. The explicit slip variant, which directly uses the handcrafted slip level, also exhibits reduced robustness under this out-of-distribution condition (e.g., \textbf{RE}$_{\text{rot}}$ of 4.241).





\bibliographystyle{ieeetr}
\bibliography{ref}

\end{document}